\documentclass[letterpaper]{article} 
\usepackage[preprint]{aaai2027}  
\usepackage[hyphens]{url}  
\usepackage{graphicx} 
\usepackage{natbib}  
\usepackage{caption} 
\usepackage{booktabs}
\usepackage{amsmath}
\usepackage{amssymb}
\usepackage{stmaryrd}
\usepackage{multirow}
\usepackage{algorithm}
\usepackage{algorithmic}
\usepackage{listings}

\lstdefinestyle{promptstyle}{
  basicstyle=\scriptsize\ttfamily,
  breaklines=true,
  breakatwhitespace=false,
  frame=single,
  framesep=3pt,
  xleftmargin=2pt,
  xrightmargin=2pt,
  showstringspaces=false,
  keepspaces=true,
  columns=fullflexible,
}

\title{SAGA: Schema-Aware Grounding for Agentic Text-to-SPARQL
       Generation}

\author{
    Yiming Zhang\textsuperscript{\rm 1},
    Koji Tsuda\textsuperscript{\rm 1,\rm 2,\rm 3}
    }
\affiliations{
    \textsuperscript{\rm 1}The University of Tokyo, Japan \\
    \textsuperscript{\rm 2}Center for Basic Research on Materials, National Institute for Materials Science, Japan \\
    \textsuperscript{\rm 3}RIKEN Center for Advanced Intelligence Project, Japan \\
}

\begin{document}

\maketitle


\begin{abstract}
Complex knowledge base question answering (KBQA) is commonly approached
through either information retrieval over a question-specific subgraph or
semantic parsing into an executable logical form.  We study the latter
paradigm.  Recent large language model agents make semantic parsing
interactive: they alternate between reasoning, querying the knowledge base,
and extending a partial SPARQL query.  This interleaving reduces reliance on
one-shot generation, but makes the quality of \emph{KB grounding} depend on
what the interaction tools expose.  Existing agents retrieve or prune
candidate properties mainly through lexical relevance and instance-level
observations, without systematically conditioning on entity types, property
domains and ranges, or the expected answer type.  We call this failure mode
\emph{type-blind grounding}.  It enlarges the grounding search space and often
produces plausible-looking but semantically incompatible triple patterns that
execute to empty results.  We propose SAGA
(\underline{S}chema-\underline{A}ware \underline{G}rounding for
\underline{A}gentic Text-to-SPARQL Generation), a training-free framework that turns property exploration into a schema-constrained grounding operation.  SAGA maintains a
persistent bidirectional type state, filters known-incompatible property
candidates at construction time, presents the remaining graph patterns in a
compact schema-annotated format, and handles missing schema information
permissively through empirical and trace-local evidence. Across nine benchmark settings over Wikidata and Freebase, SAGA achieves the highest F1 on all nine settings and the highest exact-match accuracy on eight, while reducing empty-result queries across all reported Wikidata settings.

\end{abstract}


\section{Introduction}

Knowledge base question answering (KBQA) aims to answer natural-language
questions using facts stored in a structured knowledge base.  The challenge is
especially pronounced for \emph{complex} questions, which may require
multi-hop reasoning, constrained relations, multiple subjects, or numerical
operations \cite{lan2022complexkbqa}.  Prior work distinguishes two main
approaches: \emph{information retrieval-based} (IR-based) methods build a
question-specific graph and derive answers by ranking entities or generating
from the retrieved evidence, whereas \emph{semantic parsing-based} (SP-based)
methods map the question to a symbolic logical form and execute it against the
knowledge base.  The distinction concerns the intermediate representation and
final output, not whether retrieval is used: an SP-based system may retrieve
entities, relations, and subgraphs while still producing an executable logical
form.

We focus on SP-based KBQA because an explicit query provides a verifiable
reasoning artifact and supports compositional operators such as joins,
aggregation, comparison, and Boolean queries.  Classical SP-based systems
comprise four modules---question understanding, logical parsing, KB grounding,
and KB execution \cite{lan2022complexkbqa}---where grounding aligns entities and
predicates with the target schema.  Large language models (LLMs) have recently
made these modules interactive rather than sequential.  Interactive-KBQA
\cite{xiong2024interactivekbqa} views the LLM as an agent and the KB as an
environment, alternating between a thought, a tool action, and an observation to
search nodes, retrieve graph patterns, and execute SPARQL, and SPINACH
\cite{liu2024spinach} adopts a related ReAct-style process on Wikidata.  We refer
to this family as \emph{interactive agentic semantic parsing}: logical-form
construction and KB grounding are interleaved over multiple turns.

This interaction reduces dependence on a single unsupported generation, but it
also makes the agent--KB interface a central part of the parser: at every turn
the agent can ground the next predicate only from the candidates exposed by its
tools.  Interactive-KBQA semantically ranks local graph patterns, whereas
SPINACH prunes an entity neighborhood with an additional LLM call.  For a
complex question, each added relation, constraint, or intermediate variable
enlarges the grounding search space, so a locally plausible property choice can
send the entire partial query down an incorrect branch.

We identify a systematic weakness in this grounding stage, which we call
\emph{type-blind grounding}.  Existing interactive parsers assess a candidate
property from its label, description, or observed triples, without consistently
conditioning on four pieces of symbolic state: the type of the current entity or
variable, the property's declared domain, its declared range, and the expected
answer type.  A property may therefore be lexically relevant---or observed on a
noisy instance---yet be incompatible with the semantic role assigned to the
variable, producing a syntactically valid triple pattern that executes to an
empty answer set, after which the agent spends extra turns repairing the wrong
branch.  Type blindness is thus not merely a prompting problem but a KB-grounding
problem caused by an unconstrained action space.

Existing uses of schema information do not directly solve this problem.
Schema-in-context methods place OWL, SHACL, or ShEx descriptions in the prompt
\cite{wardenga2025datashapes,kovriguina2023sparqlgen}, post-hoc validators such
as OBQC \cite{allemang2024obqc} detect violations only after a query is
generated, and instance-level constrained decoding \cite{luo2025gcr} restricts
token generation using KG connectivity but requires access to the decoder.
None of these approaches turns graph-pattern retrieval itself into a
schema-aware KB-grounding operation.

Our key idea is to use schema as a construction-time search-space constraint.
Many KGs expose entity types through instance assertions and property semantics
through domain, range, or type-constraint declarations.  Although incomplete,
this metadata provides a high-precision compatibility signal: a known domain
conflict can eliminate a property before it reaches the controller, while an
unknown property is retained rather than incorrectly rejected.  Schema thus acts
as a grounding prior, not an assumption that the KB is complete.

Building on this idea, we propose SAGA
(\underline{S}chema-\underline{A}ware \underline{G}rounding for
\underline{A}gentic Text-to-SPARQL Generation), an interactive SP-based framework that modifies the property-grounding interface while preserving executable SPARQL as the final output. SAGA maintains a persistent bidirectional type state---entity and
variable types accumulate forward from KB observations, while an expected
answer type propagates backward from the question---and at each lookup uses this
state and a KG-specific schema index to remove properties with known domain
conflicts, retain candidates whose schema is unknown, and display the remainder
with inline domain and range annotations, replacing a separate LLM pruning call.
Empirical schema induction and trace-local discoveries let the same logic
operate across Wikidata and Freebase.

We evaluate SAGA on nine benchmark settings spanning Wikidata and Freebase. Under a common execution-based protocol, SAGA achieves the highest F1 on all nine settings and the highest EM on eight. Relative to the strongest non-SAGA baseline on each benchmark, SAGA improves F1 by 2.4--20.3 points on Wikidata and by 0.2--11.2 points on Freebase. It also lowers the empty-result rate on all five Wikidata settings for which this diagnostic is reported.

Our contributions are threefold:
\begin{itemize}
  \item We position iterative text-to-SPARQL as interactive agentic semantic
        parsing and identify \emph{type-blind KB grounding} as a distinct
        failure mode at the interface between logical parsing and the KG.

  \item We introduce SAGA, a training-free schema-aware grounding framework
        that combines persistent forward and backward type signals with
        construction-time candidate filtering, compact typed graph-pattern
        observations, and permissive handling of incomplete schemas.

  \item We evaluate SAGA against comparable SP-based systems across nine
        benchmark settings on Wikidata and Freebase and show consistent gains
        in answer F1 together with lower empty-result rates and fewer LLM
        pruning calls.
\end{itemize}


\section{Related Work}

\subsection{Retrieval- and Semantic-Parsing-Based KBQA}

Complex KBQA methods are commonly grouped into information retrieval-based
(IR-based) and semantic parsing-based (SP-based) approaches
\cite{lan2022complexkbqa}.  IR-based methods retrieve question-specific
subgraphs and derive answers through ranking, graph reasoning, or generation;
recent LLM-based examples include Think-on-Graph \cite{sun2024tog} and
Reasoning on Graphs \cite{luo2024rog}.  SP-based methods instead produce
executable logical forms that support compositional operators and verifiable
execution.  Because retrieval may also serve as an intermediate grounding step
within semantic parsing, the key distinction is the output artifact.  SAGA is
SP-based, so we compare it with systems that generate executable SPARQL under
compatible protocols rather than IR-based graph-reasoning methods.

\begin{figure*}[t]
\centering
\includegraphics[width=\textwidth]{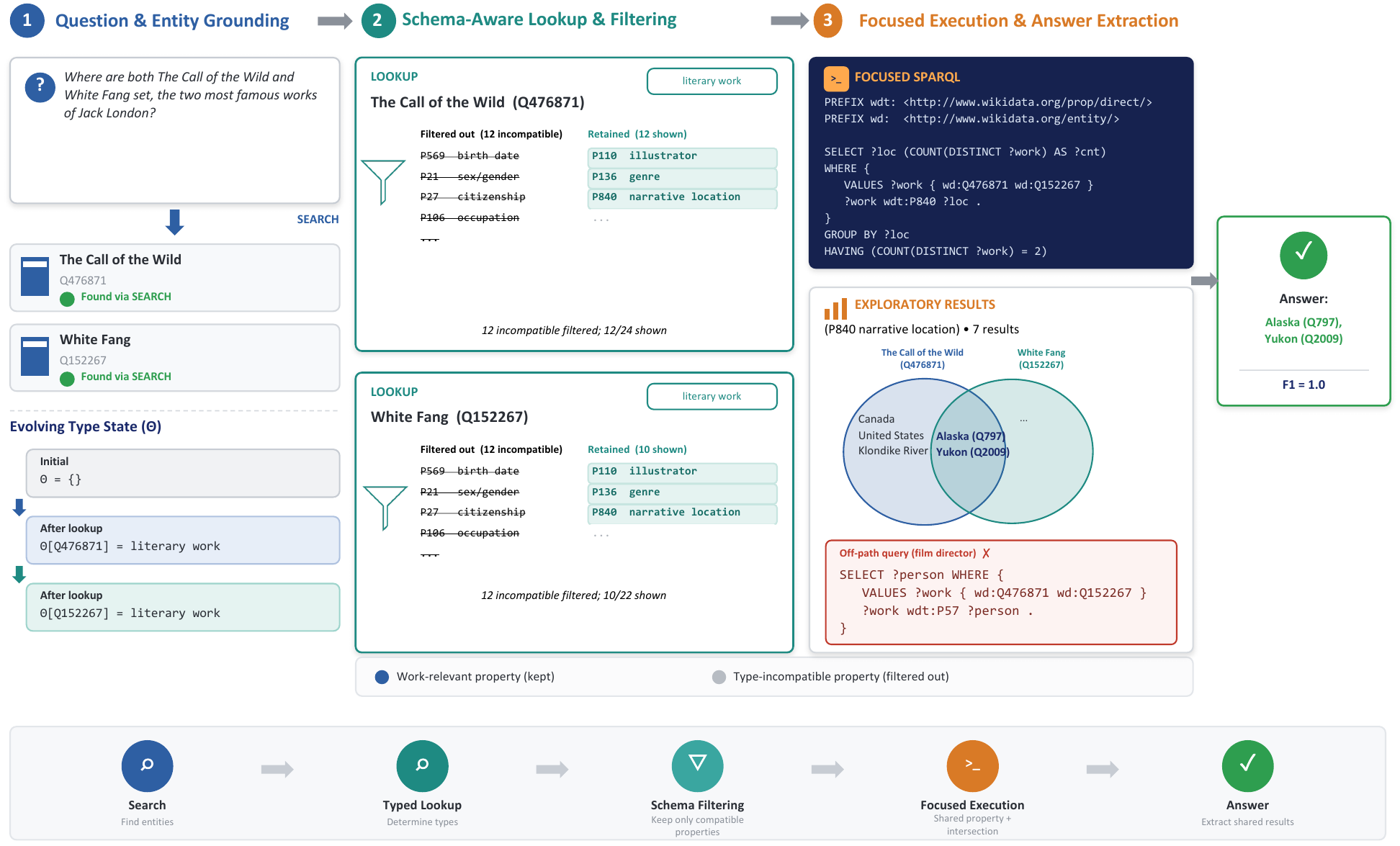}
\caption{Overview of SAGA. Instead of generating SPARQL directly from all properties of grounded entities, SAGA first performs schema-aware lookup, using inferred entity types to eliminate type-incompatible properties and retain only semantically valid relations. This substantially narrows the search space for query generation, enabling accurate multi-entity reasoning and correct intersection-based answer extraction.
}
\label{fig:overview}
\end{figure*}

\subsection{LLM-Based Interactive Semantic Parsing}

LLMs have shifted SP-based KBQA toward iterative logical-form construction.
Direct generation must resolve query structure and KG grounding in one pass,
which can yield hallucinated identifiers, incompatible predicates, or empty
results.  Interactive-KBQA \cite{xiong2024interactivekbqa} instead treats the
LLM as an agent with tools for node search, graph-pattern retrieval, and query
execution.  SPINACH \cite{liu2024spinach} similarly searches entities,
inspects neighborhoods, and revises partial queries through execution
feedback, while mKGQAgent \cite{perevalov2025text} and GRASP
\cite{walter2025grasp} provide other agentic routes to executable logical
forms.  Their grounding tools, however, select or prune predicates mainly by
lexical relevance, retrieved instances, or LLM judgments, without jointly
using entity or variable types, property domains and ranges, and the expected
answer type.  SAGA complements these controllers by making the agent--KB
grounding interface schema-aware.

\subsection{Schema-Aware Grounding and Query Validation}

Schema information has been used through prompt augmentation, constrained
generation, and post-hoc validation.  Data Shapes Prompting
\cite{wardenga2025datashapes} provides SHACL or ShEx descriptions, and
SPARQLGEN \cite{kovriguina2023sparqlgen} supplies an RDF subgraph, but the model
must still apply this context during generation.  Graph-constrained Reasoning
\cite{luo2025gcr} restricts decoding using instance-level topology, which does
not guarantee schema-level type compatibility and may require decoder access.
OBQC \cite{allemang2024obqc} checks OWL domain and range violations after a
query is formed.  SAGA instead applies schema compatibility during graph-pattern
grounding, before candidate predicates reach the controller.


\section{Method}

\subsection{Overview}
SAGA is an interactive semantic parser that helps an LLM construct an
executable SPARQL query through repeated access to a knowledge graph (KG).
Its key idea is simple: before candidate properties are shown to the LLM,
SAGA checks whether their declared domain is compatible with the type of the
entity or variable currently being explored, preventing clearly invalid
property choices while retaining candidates whose schema is missing.

The workflow has three stages.  First, SAGA builds a lightweight schema index
that stores the available domain and range information for each property.
Second, during an interaction, it maintains the types discovered for entities
and SPARQL variables, together with the expected answer type inferred from the
question.  Third, each \textsc{lookup} call uses this information to filter and
annotate the local neighborhood before returning it to the LLM, which then
continues constructing, executing, and revising the query as in prior agentic
KBQA systems.  Thus, SAGA changes the agent--KG interface rather than the LLM or
its token-level decoding process.
Figure~\ref{fig:overview} illustrates this pipeline using the question
``Where are both \textit{The Call of the Wild} and \textit{White Fang} set?'';
the three lower panels correspond to the schema index, lookup-time
type-constrained grounding, and persistent type state described in the
following subsections.

\subsection{Problem Setup and Interactive Query Construction}

We represent a knowledge graph (KG) as
$\mathcal{G}=(\mathcal{E},\mathcal{P},\mathcal{T})$, where $\mathcal{E}$ is
the set of entities (e.g., Wikidata's \texttt{Q212} for the United States),
$\mathcal{P}$ is the set of \emph{properties}---the labeled edges that
connect a subject to an object in a triple (e.g., \texttt{P17}, ``country'')---
and $\mathcal{T}$ is the set of subject--property--object triples.  Triple
objects may be entities or literal values such as numbers or dates. 

Every entity belongs to zero or more \emph{types} (e.g., \emph{human} or
\emph{sovereign state}).  We write $\mathcal{C}$ for the set of all types and
$\tau(e)\subseteq\mathcal{C}$ for the types declared for entity $e$ (e.g.,
$\tau(\texttt{Q212})=\{\text{sovereign state}\}$).  Each property $p$ may
additionally declare a \emph{domain} $\delta(p)\subseteq\mathcal{C}$ and a
\emph{range} $\rho(p)\subseteq\mathcal{C}$: the domain lists the types an
entity must have in order to appear as the \emph{subject} of a $p$-triple,
and the range lists the types required of the \emph{object}.  For instance,
$\delta(\texttt{P607})=\{\text{character, battle, sabotage}\}$ means only
entities of those types are valid subjects of a \texttt{P607} (``conflict'')
triple.  Either $\delta(p)$ or $\rho(p)$ may be empty when the KG does not
declare this information for $p$; the \emph{Schema Index and Type State}
subsection below explains how SAGA handles that case.

Given a natural-language question $q$, the system constructs a SPARQL query
$\sigma$.  Executing the final query on the KG produces the predicted answer
set $\hat{A}_q$:
\begin{equation}
  \hat{A}_q=\llbracket\sigma\rrbracket_{\mathcal{G}},
  \label{eq:query-execution}
\end{equation}
where $\llbracket\sigma\rrbracket_{\mathcal{G}}$ denotes the result of
executing $\sigma$ on $\mathcal{G}$, i.e., the set of entities or literals
returned by the SPARQL endpoint.

Following Interactive-KBQA \cite{xiong2024interactivekbqa}, an LLM
\emph{controller} builds the query through a sequence of tool calls.  The
controller observes the question and the previous
actions and observations, then chooses one of five actions: \textsc{search}
for retrieving entities from a text query, \textsc{lookup} for inspecting the
neighborhood of an entity, \textsc{examples} for viewing example uses of a
property, \textsc{execute} for running a partial or complete SPARQL query,
and \textsc{stop} for ending the interaction.

The main grounding bottleneck is \textsc{lookup}.  For an entity $e$, its raw
outgoing neighborhood is
\begin{equation}
  \mathcal{N}(e)=\{(p,v)\mid(e,p,v)\in\mathcal{T}\},
  \label{eq:raw-neighborhood}
\end{equation}
where $p$ is a property and $v$ is the corresponding object value; that is,
$\mathcal{N}(e)$ simply collects every (property, value) pair reachable by a
single outgoing edge from $e$---the candidate triples that \textsc{lookup}
would hand to the controller.  A large entity may expose hundreds of such
properties.  Existing interactive parsers mainly prune them using lexical
relevance or observed instances, without systematically checking whether a
property's schema is compatible with the current type.  SAGA performs this
check before the candidates are presented to the controller.

\subsection{Schema Index and Type State}

\paragraph{Schema index.}
Before inference, SAGA creates one metadata entry per property, recording its
known domain $\delta(p)$ and range $\rho(p)$ (defined above) so that they can
be looked up without querying the KG at run time.  For Wikidata, formal
entries come from P2302 type constraints; for Freebase, they come from the
curated \texttt{fb\_roles} schema.  When a property lacks a formal
declaration, SAGA instead samples several subject--object pairs from the KG
and stores the most frequently observed subject and object types as soft
domain and range evidence.  A property for which neither source is available
is left unconstrained ($\delta(p)=\rho(p)=\emptyset$) rather than removed
from the index.

\paragraph{Persistent type state.}
During query construction, SAGA maintains a \emph{type state} $\Theta$: a
running record that maps every entity or SPARQL variable encountered so far
to the types currently known for it (formally, $\Theta[x]\subseteq\mathcal{C}$
for entity or variable $x$).  Looking up an entity adds its declared types
$\tau(e)$ to the state.  Likewise, an explicit type pattern in a partial
query, such as \texttt{?x wdt:P31 wd:Q5}, records that the variable
\texttt{?x} is a human.  Because $\Theta$ persists across the whole
interaction, information discovered in an early step can guide property
selection later in the same interaction---this is what we mean by
\emph{persistent}.

Before the first tool call, the controller also predicts the \emph{expected
answer type} (Supplementary Appendix~F.1): the type the final answer is
expected to belong to (e.g., for
``Where are both \textit{The Call of the Wild} and \textit{White Fang} set?'', the expected answer type is \emph{location}).
This expected type is used only as guidance, not as a hard filter: it can
increase the priority of properties whose range matches it, but an
intermediate variable may legitimately have a different type from the final
answer, so it is never used to discard a property outright.  SAGA additionally
retains execution outcomes and newly observed type evidence in the
interaction history so that the controller does not repeatedly make the same
failed choice.

\subsection{Type-Constrained Grounding}

When \textsc{lookup}$(e)$ is called, SAGA retrieves the types $\tau(e)$ of
$e$ (introduced above under \emph{Persistent type state}) and the set $P(e)$
of properties occurring in $e$'s raw neighborhood $\mathcal{N}(e)$.  A property
$p\in P(e)$ survives the filter unless it has a declared domain that
conflicts with every one of $e$'s known types:
\begin{equation}
  P_{\tau}(e)=\{p\in P(e)\mid
  \delta(p)=\emptyset\ \lor\
  \delta(p)\cap\tau(e)\neq\emptyset\},
  \label{eq:type-compatible-properties}
\end{equation}
Intuitively, Eq.~\eqref{eq:type-compatible-properties} keeps a property $p$
in one of two situations: either $p$'s domain is unknown ($\delta(p)=\emptyset$,
so there is no basis for rejecting it), or $p$'s domain and $e$'s types
overlap ($e$ is a plausible subject for $p$).  The only remaining case---a
\emph{known} domain that shares no type with $\tau(e)$---is a demonstrated
type conflict, and it is the sole reason a property is removed.  $P_\tau(e)$
denotes the properties that survive this check, and every pair $(p,v)\in
\mathcal{N}(e)$ whose property is not in $P_\tau(e)$ is dropped from the
observation returned to the controller.  Consequently, a known domain
conflict is rejected before generation, whereas missing schema information
never causes rejection.  When no type is known for $e$ (i.e., $\tau(e)=\emptyset$),
Eq.~\eqref{eq:type-compatible-properties} would incorrectly reject every
property that \emph{does} have a declared domain, since intersecting any
non-empty set with $\emptyset$ is always empty; SAGA therefore skips the
filter in this case and falls back to the same LLM-based neighborhood pruning
used by the underlying interactive parser.

Figure~\ref{fig:overview} shows that this filtering is entity-type-dependent
rather than property-specific: the same property can be retained for an entity
of a compatible type and removed for one that is not.  In the illustrated
example, SAGA processes both literary-work entities Q476871
(\textit{The Call of the Wild}) and Q152267 (\textit{White Fang})
symmetrically: 12 of 24 and 12 of 22 properties are removed from each---those
whose declared domain (e.g., \emph{human} or \emph{country}) conflicts with
\emph{literary work}---while work-compatible properties such as
\texttt{P136}~(\textit{genre}) are retained with inline domain annotations.
The same property \texttt{P136} would be removed for a human entity, since its
declared domain does not include \emph{human}.  Recognizing both novels as
the same type also enables the controller to construct a unified multi-entity
SPARQL using \texttt{VALUES} and a \texttt{GROUP BY}/\texttt{HAVING}
intersection to retrieve their shared narrative settings.

For each retained property, SAGA returns its identifier, label, short
description, and available domain and range information, rendered in a
compact entry of the form:
\smallskip
\begin{small}
\texttt{label (PID): description}\\
\texttt{~~domain: type names}\\
\texttt{~~range: type names}
\end{small}
\smallskip
This representation makes the schema evidence visible to the controller at
the moment of property selection.  Because filtering and annotation are
performed by the \textsc{lookup} tool itself, SAGA does not require a
separate LLM call to prune the same neighborhood.

SAGA also accumulates schema evidence during execution.  When a property with
no indexed range returns non-empty results, the system inspects the types of a
small sample of returned entities and records them as trace-local evidence.
This evidence can guide later steps of the current interaction, but it does not
overwrite the offline schema index or act as a hard constraint.

\subsection{Agentic Loop}

At each step, the controller selects one of the five actions described
above.  A \textsc{lookup}$(e)$ call runs the type-retrieval-and-filtering
process of Eq.~\eqref{eq:type-compatible-properties}; an
\textsc{execute}$(\sigma)$ call runs the (partial) query and updates the
persistent type state $\Theta$ with any newly observed types; and
\textsc{stop} ends the interaction and returns the final query $\sigma^{*}$.
Every action and its resulting observation are appended to the interaction
history $H$, which the controller conditions on at the next step.
Algorithm~\ref{alg:saga-main} formalizes this loop: $H$ contains the
action--observation history, $a$ and $o$ denote the current action and
observation, $\Theta$ is the persistent type state, and $\sigma^{*}$ is the
final query returned by the controller.  \textsc{TypeFilter} and
\textsc{CompactFormat} implement the type check and compact neighborhood
rendering described in the \emph{Type-Constrained Grounding} subsection, while
\textsc{LLMPrune} is the fallback used when no entity type is available.  The
controller and answer-type prompts are provided in Supplementary Appendix~F.
SAGA operates at the
tool-call level rather than during token decoding, so it is compatible with
both open and closed LLMs and requires no access to model internals; unlike
instance-level constrained decoding \cite{luo2025gcr}, its main constraints
come from KG schema and can remain useful when local instance evidence is
sparse.

\begin{algorithm}[h]
\caption{SAGA agent loop}
\label{alg:saga-main}
\textbf{Input}: question $q$, knowledge graph $\mathcal{G}$\\
\textbf{Output}: final SPARQL query $\sigma^{*}$
\begin{algorithmic}[1]
\STATE $H\leftarrow[]$; $\Theta\leftarrow\{\}$
\STATE $\textit{answer\_type}\leftarrow\text{InferAnswerType}(q)$
\STATE $\textit{stopped}\leftarrow\textbf{false}$
\WHILE{not $\textit{stopped}$}
  \STATE $a\leftarrow\text{Controller}(q,H,\Theta,\textit{answer\_type})$
  \IF{$a=\textsc{lookup}(e)$}
    \STATE $\textit{types}\leftarrow\text{GetTypes}(e,\mathcal{G})$
    \STATE $\textit{neighbors}\leftarrow\text{GetNeighborhood}(e,\mathcal{G})$
    \STATE $\Theta[e]\leftarrow\textit{types}$
    \IF{$\textit{types}\neq\emptyset$}
      \STATE $\textit{kept}\leftarrow\text{TypeFilter}(\textit{neighbors},\textit{types})$
      \STATE $o\leftarrow\text{CompactFormat}(\textit{kept},\textit{answer\_type})$
    \ELSE
      \STATE $o\leftarrow\text{LLMPrune}(\textit{neighbors},q)$
    \ENDIF
  \ELSIF{$a=\textsc{execute}(\sigma)$}
    \STATE $o\leftarrow\text{Execute}(\sigma,\mathcal{G})$
    \STATE $\Theta\leftarrow\text{UpdateTypeState}(\Theta,\sigma,o)$
  \ELSIF{$a=\textsc{stop}$}
    \STATE $o\leftarrow\emptyset$; $\textit{stopped}\leftarrow\textbf{true}$
  \ELSE
    \STATE $o\leftarrow\text{ExecuteAction}(a)$
  \ENDIF
  \STATE $H.\text{append}((a,o))$
\ENDWHILE
\STATE $\sigma^{*}\leftarrow\text{FinalQuery}(H)$
\STATE \textbf{return} $\sigma^{*}$
\end{algorithmic}
\end{algorithm}

\section{Experiments}

\paragraph{Datasets.}
We evaluate SAGA on nine benchmark settings over Wikidata and Freebase.
For Wikidata, QALD-7 \cite{usbeck20177th} contains 45 test
questions, QALD-9-plus \cite{perevalov2022qald9plus} provides a
127-question Wikidata test split, and QALD-10 \cite{usbeck2023qald10}
contains 383 test questions. WikiWebQuestions (WWQ) is the Wikidata
adaptation of WebQuestionSP \cite{yih2016webqsp}, with 450 development and
1{,}409 test questions. The SPINACH dataset \cite{liu2024spinach} contains
149 development and 156 test questions collected from complex in-the-wild
questions. LC-QuAD 2.0 \cite{dubey2019lcquad2} contributes 826 Wikidata test
questions, including multi-hop and Boolean queries.

For Freebase, WebQSP \cite{yih2016value} contains 1{,}628 test questions
with annotated topic entities and SPARQL queries. ComplexWebQuestions (CWQ)
\cite{talmor2018web} contains 3{,}273 compositional test questions involving
conjunctions, comparisons, and superlatives. GrailQA \cite{gu2021beyond}
contains 1{,}000 test questions covering i.i.d., compositional, and zero-shot
generalization settings.

\paragraph{Baselines.}
We compare SAGA with Zero-shot SPARQL, Entity Linking
\cite{lan2022complexkbqa}, Data Shapes Prompting
\cite{wardenga2025datashapes}, OBQC Repair
\cite{allemang2024obqc}, mKGQAgent \cite{perevalov2025text}, GRASP
\cite{walter2025grasp}, Interactive-KBQA \cite{xiong2024interactivekbqa} and SPINACH \cite{liu2024spinach}. These baselines produce executable SPARQL queries and are evaluated using
execution-based metrics. SPINACH shares SAGA's multi-turn process of entity
search, property inspection, query construction, and execution, while SAGA
adds schema-aware grounding to the interaction process.

\paragraph{Metrics.}
Let $\hat{A}$ be the predicted and $A^*$ the gold answer set for a question.  We
report two execution-based metrics averaged over all questions.
\emph{Macro-averaged F1} is the harmonic mean of set precision and recall:
\begin{equation}
  \mathrm{F1} = \frac{1}{|Q|}\sum_{q \in Q}
    \frac{2\,|\hat{A}_q \cap A^*_q|}{|\hat{A}_q| + |A^*_q|},
\end{equation}
where $|Q|$ is the number of test questions and an empty prediction yields
$\mathrm{F1}_q = 0$.

\emph{Exact Match (EM)} is the fraction of questions for which the prediction
is a perfect set match with the gold:
\begin{equation}
  \mathrm{EM} = \frac{1}{|Q|}\sum_{q \in Q}
    \mathbf{1}\!\left[\hat{A}_q = A^*_q\right]
  = \frac{1}{|Q|}\sum_{q \in Q} \mathbf{1}\!\left[\mathrm{F1}_q = 1\right].
\end{equation}

\begin{table*}[t]
\centering
\caption{F1 / EM (\%) on Wikidata KGQA benchmarks. EM is the fraction of
questions with an exact answer-set match.}
\label{tab:main}
\small
\setlength{\tabcolsep}{3pt}
\begin{tabular}{l*{6}{cc}}
\toprule
& \multicolumn{2}{c}{QALD-7}
& \multicolumn{2}{c}{QALD-9+}
& \multicolumn{2}{c}{QALD-10}
& \multicolumn{2}{c}{WWQ-test}
& \multicolumn{2}{c}{SPINACH-test}
& \multicolumn{2}{c}{LC-QuAD~2.0} \\
\cmidrule(lr){2-3}\cmidrule(lr){4-5}\cmidrule(lr){6-7}
\cmidrule(lr){8-9}\cmidrule(lr){10-11}\cmidrule(lr){12-13}
Method & F1 & EM & F1 & EM & F1 & EM & F1 & EM & F1 & EM & F1 & EM \\
\midrule
Zero-shot SPARQL
  & 18.58 & 18.00 & 15.11 & 12.50 & 16.54 & 15.74
  & 18.06 & 13.28 &  3.71 &  1.21 &  3.86 &  3.39 \\
Entity Linking \cite{lan2022complexkbqa}
  & 25.08 & 24.44 & 21.72 & 17.97 & 25.89 & 24.28
  & 18.34 & 13.48 &  3.90 &  1.30 &  5.74 &  4.84 \\
Data Shapes Prompting \cite{wardenga2025datashapes}
  & 42.19 & 35.56 & 39.40 & 31.25 & 31.76 & 30.03
  & 18.27 & 13.41 &  4.19 &  1.95 &  9.83 &  8.22 \\
OBQC Repair \cite{allemang2024obqc}
  & 31.49 & 26.67 & 28.79 & 24.22 & 33.88 & 31.85
  & 17.27 & 11.51 &  4.66 &  2.05 &  9.34 &  7.50 \\
mKGQAgent \cite{perevalov2025text}
  & 33.88 & 28.89 & 31.94 & 26.56 & 39.18 & 35.51
  & 31.18 & 22.92 &  3.03 &  1.30 & 15.15 & 13.42 \\
GRASP \cite{walter2025grasp}
  & 41.24 & 33.78 & 42.21 & 35.75 & 43.95 & 41.70
  & 33.69 & 24.90 & 18.02 & \textbf{10.30} & 20.48 & 17.95 \\
Interactive-KBQA \cite{xiong2024interactivekbqa}
  & 18.96  & 13.33 & 24.38 & 17.97 & 37.10 & 32.64
  & 29.89 & 21.45 & 5.22 & 1.29 & 19.93 & 15.72 \\
SPINACH \cite{liu2024spinach}
  & 44.56 & 37.78 & 43.54 & 36.65 & 41.64 & 39.69
  & 34.36 & 32.21 & 17.00 &  8.33 & 25.83 & 23.58 \\
\midrule
SAGA (LLaMA~3.1-70B)
  & 49.40 & 44.44
  & 40.02 & 33.59
  & 42.41 & 40.21
  & 38.28 & 31.94
  & 14.85 &  3.23
  & 26.99 & 24.48 \\
SAGA (gpt-oss-120B)
  & \textbf{64.89} & \textbf{53.33}
  & \textbf{55.66} & \textbf{48.44}
  & \textbf{53.04} & \textbf{48.04}
  & \textbf{51.49} & \textbf{45.56}
  & \textbf{20.60} & 6.45
  & \textbf{28.25} & \textbf{25.39} \\
\bottomrule
\end{tabular}
\end{table*}

\begin{table*}[htbp]
\centering
\caption{F1 / EM (\%) on Freebase KGQA benchmarks.}
\label{tab:freebase}
\setlength{\tabcolsep}{3.5pt}
\begin{tabular}{l*{3}{cc}}
\toprule
& \multicolumn{2}{c}{WebQSP}
& \multicolumn{2}{c}{CWQ}
& \multicolumn{2}{c}{GrailQA} \\
\cmidrule(lr){2-3}\cmidrule(lr){4-5}\cmidrule(lr){6-7}
Method & F1 & EM & F1 & EM & F1 & EM \\
\midrule
Zero-shot SPARQL
  &  11.40 & 8.06 &  6.05 & 3.00 & 11.46 & 7.07 \\
Entity Linking \cite{lan2022complexkbqa}
  &  13.14 & 9.00 &  7.52 & 4.10 & 14.47 & 8.10 \\
Data Shapes Prompting \cite{wardenga2025datashapes}
  &  15.04 & 9.05 &  8.26 & 4.00 & 16.45 & 9.10 \\
OBQC Repair \cite{allemang2024obqc}
  &  19.49 & 13.20 &  10.55 & 9.00 & 15.38 & 7.85 \\
mKGQAgent \cite{perevalov2025text}
  & 31.78 & 26.19 & 11.94 & 9.76 & 40.28 & 34.51 \\
GRASP \cite{walter2025grasp}
  & 30.24 & 24.78 & 12.21 & 8.75 & 43.05 & 40.60 \\
Interactive-KBQA \cite{xiong2024interactivekbqa}
  & 22.50 & 15.85 & 7.90 & 4.98  & 17.55 & 15.10 \\
SPINACH \cite{liu2024spinach}
  & 31.83 & 29.10 &  15.27 &  11.38 & 46.95 & 44.20 \\
\midrule
SAGA (LLaMA~3.1-70B)
  & 28.84 & 24.02
  & 13.42 & 10.88
  & 36.55 & 33.70 \\
SAGA (gpt-oss-120B)
  & \textbf{40.10} & \textbf{35.69}
  & \textbf{15.47} & \textbf{13.81}
  & \textbf{58.13} & \textbf{55.60} \\
\bottomrule
\end{tabular}
\end{table*}

We additionally report the \emph{empty-result rate}, the proportion of
generated queries that return no answer from the SPARQL endpoint.

\subsection{Implementation Details}

All experiments are conducted with a 120B-parameter open-weight instruction-tuned
LLM (gpt-oss-120B \cite{openai2025gptoss120bgptoss20bmodel}) served through a local vLLM endpoint.  We use greedy decoding
(temperature~$= 0$) and cap each response at 2{,}048 tokens to obtain
deterministic outputs.  SAGA is additionally evaluated with LLaMA-3.1-70B \cite{grattafiori2024llama} across
all nine benchmark settings under the same decoding configuration.  For all
baseline systems, we use gpt-oss-120B with identical inference settings so that
differences in results reflect only the grounding strategy.

\section{Results}

\subsection{Results on Wikidata}

Table~\ref{tab:main} reports execution-based F1 and EM on six Wikidata
benchmarks.  SAGA with gpt-oss-120B achieves the highest F1 on all six
datasets.

Relative to SPINACH, SAGA with gpt-oss-120B improves F1 by 2.4--20.3 points
across the six datasets (Table~\ref{tab:main}), with the largest gains on
QALD-7 and WWQ-test, where query construction depends strongly on selecting
valid properties over multiple turns.  The gap between SAGA and SPINACH indicates
that exposing schema constraints during property grounding is more effective
than an otherwise similar multi-turn interaction without the type-aware
interface.  A manual error analysis on QALD-7 and QALD-9-plus confirms that
36.9\% of SPINACH failures (41 of 111) are caused by type-blindness---selecting
a property whose declared domain conflicts with the entity type---which SAGA
corrects through construction-time filtering (Supplementary Appendix~A).  SAGA achieves these
gains largely without increasing the interaction budget: across the four
reported datasets it uses the same number of actions as SPINACH on QALD-7,
fewer on QALD-9-plus and SPINACH-test (a 29\% reduction on SPINACH-test), and
slightly more on QALD-10 (Supplementary Appendix~B).

Table~\ref{tab:lcquad} separates LC-QuAD 2.0 into Boolean and factoid questions.
SAGA raises Boolean F1 from 25.6 to 55.6, a gain of 30.0 points, while its
factoid F1 is 1.0 point lower than SPINACH.  The Boolean improvement is large
enough to raise overall F1 from 25.83 to 28.25 and EM from 23.58 to 25.39.
Thus, the overall LC-QuAD gain is concentrated in questions whose logical form
requires a Boolean answer rather than an answer set.

\begin{table}[t]
\centering
\caption{F1 / EM (\%) on LC-QuAD 2.0 by question type.}
\label{tab:lcquad}
\setlength{\tabcolsep}{4pt}
\small
\begin{tabular}{l*{3}{cc}}
\toprule
& \multicolumn{2}{c}{Boolean}
& \multicolumn{2}{c}{Factoid}
& \multicolumn{2}{c}{Overall} \\
\cmidrule(lr){2-3}\cmidrule(lr){4-5}\cmidrule(lr){6-7}
Method & F1 & EM & F1 & EM & F1 & EM \\
\midrule
SPINACH
  & 25.6 & 25.6 & \textbf{25.9} & \textbf{23.3} & 25.83 & 23.58 \\
SAGA (ours)
  & \textbf{55.6} & \textbf{55.6} & 24.9 & 21.7
  & \textbf{28.25} & \textbf{25.39} \\
\bottomrule
\end{tabular}
\end{table}

The empty-result analysis in Table~\ref{tab:empty} provides complementary
evidence: SAGA lowers the empty-result rate on every reported setting, most
notably on WWQ-test (75.5\% to 27.0\%), QALD-9-plus (38.6\% to 25.8\%), and
QALD-10 (32.4\% to 21.9\%).  This is consistent with the intended role of
schema-aware grounding, in which properties with known type conflicts are
removed before they can enter a partial query.

\begin{table}[t]
\centering
\caption{Empty-result rate (\%) on Wikidata benchmarks. Lower is better.}
\label{tab:empty}
\setlength{\tabcolsep}{3pt}
\resizebox{\columnwidth}{!}{
\begin{tabular}{lccccc}
\toprule
Method & QALD-7 & QALD-9+ & QALD-10 & WWQ-test & SPINACH-test \\
\midrule
SPINACH & 22.2 & 38.6 & 32.4 & 75.5 & 52.4 \\
SAGA & \textbf{15.6} & \textbf{25.8} & \textbf{21.9}
     & \textbf{27.0} & \textbf{49.3} \\
\bottomrule
\end{tabular}
}
\end{table}

\subsection{Results on Freebase}

Table~\ref{tab:freebase} reports F1 and EM on WebQSP, CWQ, and GrailQA.
Direct SPARQL generation remains weak on all three datasets, while the
multi-turn SPINACH baseline performs substantially better, showing the
importance of interacting with the KG during logical-form construction.  SAGA
further improves over SPINACH by 0.2--11.2 F1 across the three datasets
(Table~\ref{tab:freebase}), though the margin on CWQ is small.  These
improvements show that schema-aware property grounding transfers beyond
Wikidata and remains effective under a different KG schema and relation
inventory.

\subsection{Ablation Study}

Table~\ref{tab:ablation} evaluates the three grounding mechanisms defined in
the ablation protocol on the five Wikidata benchmarks with complete ablation
results, reporting the unweighted macro average across them.  Full SAGA
obtains the highest average F1 of 49.14.  Removing persistent type state
causes the largest average decrease (5.43 points), followed by removing
type-constrained filtering (4.97 points), showing that both construction-time
filtering and cross-turn type memory contribute substantially.

\begin{table}[t]
\centering
\caption{Component ablation on Wikidata. Values are F1 (\%).}
\label{tab:ablation}
\resizebox{\columnwidth}{!}{
\begin{tabular}{lrrrrr}
\toprule
Variant & QALD-7 & QALD-9+ & QALD-10 & WWQ-test & SPINACH-test \\
\midrule
Full SAGA
  & 64.89 & 55.66 & 53.04 & 51.49 & 20.60\\
$-$ Type-constrained filtering
  & 58.05 & 49.97 & 51.19 & 48.97 & 12.64 \\
$-$ Persistent type state
  & 53.37 & 48.58 & 50.73 & 50.12 & 15.71 \\
$-$ Answer-type prior
  & 63.14 & 53.87 & 50.43 & 50.60 & 16.08\\
\bottomrule
\end{tabular}
}
\end{table}

The contribution is not uniform across datasets.  Type filtering is
particularly important on QALD-7, QALD-9-plus, and SPINACH-test, where its
removal lowers F1 by 6.84, 5.69, and 7.96 points; persistent type state has the
strongest effect on QALD-7 and QALD-9-plus, with drops of 11.52 and 7.08 points.
The answer-type prior has the smallest average effect of the three mechanisms:
removing it lowers F1 on all five reported datasets, with the largest drop on
SPINACH-test (4.52 points) and the smallest on WWQ-test (0.89 points).  This is
consistent with the prior's role in biasing the controller toward Boolean
(ASK) queries before any entity lookup (Supplementary Appendix~D), while Full SAGA still
achieves the best macro average.

\subsection{Robustness to Incomplete Schema}
\label{sec:coverage}

Table~\ref{tab:coverage} evaluates SAGA after retaining different fractions of
Wikidata P2302 domain constraints.  Full coverage obtains the highest average
F1 (48.55) and the best result on QALD-7, QALD-9-plus, and SPINACH-test.  At
0\% coverage the average falls to 46.08, with drops of 8.92 points on QALD-7 and
2.35 on QALD-9-plus, supporting the usefulness of formal schema constraints.

\begin{table}[t]
\centering
\caption{F1 (\%) under different P2302 schema-coverage levels.}
\label{tab:coverage}
\setlength{\tabcolsep}{4pt}
\resizebox{\columnwidth}{!}{
\begin{tabular}{lccccc}
\toprule
Coverage & QALD-7 & QALD-9+ & QALD-10 & SPINACH-test & Avg \\
\midrule
0\%   & 55.97 & 53.31 & 57.20 & 17.85 & 46.08 \\
25\%  & 64.78 & 47.81 & \textbf{57.78} & 17.35 & 46.93 \\
50\%  & 59.49 & 49.78 & 49.71 & 18.00 & 44.25 \\
75\%  & 53.99 & 52.81 & 54.72 & 13.82 & 43.84 \\
100\% & \textbf{64.89} & \textbf{55.66} & 53.04
      & \textbf{20.60} & \textbf{48.55} \\
\bottomrule
\end{tabular}
}
\end{table}

The trend is not monotonic: QALD-10 performs best at 25\% coverage, and several
intermediate masks underperform both endpoints, showing that robustness depends
not only on the amount of schema but also on which constraints are retained.
Importantly, SAGA degrades gracefully when schema is absent because unknown
properties remain available rather than being rejected.  The best average at
100\% coverage indicates that the complete schema is beneficial in aggregate,
while the dataset-level exceptions motivate more selective or confidence-weighted
constraint use in future work.  Detailed qualitative examples and case studies,
including both successful recoveries and a representative failure, are
collected in the supplementary material.


\section{Conclusion}

We introduced SAGA, a schema-aware grounding framework for agentic SPARQL
construction.  SAGA maintains persistent entity and variable types together
with an expected answer type, filters predicates with known domain conflicts,
and annotates retained graph patterns with domain and range information while
preserving candidates with incomplete schema through permissive fallbacks.
Across Wikidata and Freebase benchmarks, SAGA improves answer F1, reduces
empty-result queries, and removes a separate LLM pruning call.  These results
show that KG schema should shape the LLM's grounding action space during
logical-form construction, rather than serving only as prompt context or a
post-hoc validator.  Its main limitation is dependence on explicit type
annotations, motivating richer type inference for incompletely typed entities.


\newpage
\bibliography{aaai2027}

\clearpage
\appendix
\setcounter{secnumdepth}{2}
\section*{Supplementary Material}

\section{SPINACH Error Distribution}
\label{app:errors}

Table~\ref{tab:errors} breaks down SPINACH's failures on QALD-7 and
QALD-9-plus into three mutually exclusive categories.
\emph{Type blindness} covers cases where the generated SPARQL applies a
property whose declared domain is incompatible with the subject entity's type
(e.g., applying a relation whose domain is \textit{taxon} to a
\textit{human} entity).
\emph{Entity linking} covers cases where the entity search returns a wrong
QID that no type check could correct (e.g., the correct entity does not appear
in the candidate list).
\emph{Structure error} covers cases where the entity and properties are
correct but the SPARQL form is wrong (e.g., a missing path expression or
wrong aggregation).

\begin{table}[h]
\centering
\caption{SPINACH error categories on QALD-7 and QALD-9-plus test sets.
Numbers in parentheses are fractions of the wrong-answer subset.}
\label{tab:errors}
\setlength{\tabcolsep}{5pt}
\small
\begin{tabular}{lrrrr}
\toprule
Dataset & Wrong & Type blind. & Entity link. & Structure \\
\midrule
QALD-7   & 28 & 18 (64.3\%) &  2 (7.1\%) &  8 (28.6\%) \\
QALD-9+  & 83 & 23 (27.7\%) & 35 (42.2\%) & 25 (30.1\%) \\
\midrule
Combined & 111 & 41 (36.9\%) & 37 (33.3\%) & 33 (29.7\%) \\
\bottomrule
\end{tabular}
\end{table}

Type blindness is the single largest error category on QALD-7, accounting
for 64.3\% of failures.  On QALD-9-plus the picture is more balanced: entity
linking errors are slightly more frequent (42.2\%), though SAGA's
type-constrained grounding also indirectly helps entity linking by signalling
to the LLM that a resolved entity is the wrong type (see Appendix~\ref{app:wikileaks-case}).
Across both datasets, 36.9\% of failures are type-blindness errors that
SAGA's construction-time filtering is designed to address.


\section{Agent Action Efficiency}
\label{app:efficiency}

Table~\ref{tab:efficiency} compares the average number of agent actions per
question between SPINACH and SAGA (both using gpt-oss-120B).  SAGA uses the
same number of total actions on QALD-7, fewer on QALD-9-plus and
SPINACH-test, and slightly more on QALD-10.  The average lookup calls per
question increase slightly on QALD-7 and QALD-10 (SAGA retries entity lookups
more often when a type mismatch signals the wrong entity), but decrease on
SPINACH-test where SAGA's tighter grounding reduces dead-end exploration.
Overall, schema-aware grounding adds no meaningful interaction overhead.

\begin{table}[h]
\centering
\caption{Average agent actions and entity lookup calls per question.
Both systems use gpt-oss-120B; F1 is shown for reference.}
\label{tab:efficiency}
\setlength{\tabcolsep}{4pt}
\small
\begin{tabular}{llrrr}
\toprule
Dataset & System & F1 & Avg.\ actions & Avg.\ lookups \\
\midrule
\multirow{2}{*}{QALD-7}
  & SPINACH & 44.6 & 5.47 & 0.67 \\
  & SAGA    & \textbf{64.9} & \textbf{5.47} & 0.89 \\
\midrule
\multirow{2}{*}{QALD-9+}
  & SPINACH & 43.5 & 6.60 & 0.94 \\
  & SAGA    & \textbf{55.7} & \textbf{5.88} & 0.86 \\
\midrule
\multirow{2}{*}{QALD-10}
  & SPINACH & 41.6 & 5.98 & 0.91 \\
  & SAGA    & \textbf{53.0} & 6.25 & 1.14 \\
\midrule
\multirow{2}{*}{SP-test}
  & SPINACH & 17.0 & 7.71 & 0.60 \\
  & SAGA    & \textbf{20.6} & \textbf{5.49} & 0.39 \\
\bottomrule
\end{tabular}
\end{table}


\section{Neighborhood Filtering Statistics}
\label{app:filtering}

Table~\ref{tab:filtering} reports neighborhood filtering statistics measured
across the SAGA evaluation sets that contain gold SPARQL annotations.  For
each dataset, \emph{Edge reduction} is the fraction of candidate
property-value pairs removed by type-constrained filtering.
\emph{Gold-prop recall} is the fraction of gold SPARQL properties that survive
the filter (i.e., still appear in the schema-compact neighborhood shown to the
LLM).

\begin{table}[h]
\centering
\caption{Neighborhood reduction and gold-property recall per dataset.
Lower edge reduction means fewer candidates removed;
higher gold-prop recall means the filter is more conservative.}
\label{tab:filtering}
\setlength{\tabcolsep}{5pt}
\small
\begin{tabular}{lrrr}
\toprule
Dataset & Raw edges & Edge reduc. & Gold-prop recall \\
\midrule
QALD-9+      & 2{,}718  & 26.6\% & 61.4\% \\
QALD-10      & 12{,}352 & 28.6\% & 59.7\% \\
SPINACH-test & 3{,}647  & 22.4\% & 64.8\% \\
LC-QuAD~2.0  & 21{,}034 & 25.5\% & 62.6\% \\
\midrule
Avg.         & ---      & 25.8\% & 62.1\% \\
\bottomrule
\end{tabular}
\end{table}

On average, SAGA removes 25.8\% of raw candidate edges per entity lookup
while retaining 62.1\% of the gold properties needed to answer each question.
The 37.9\% of gold properties that do not survive the filter represent cases
where the Wikidata P2302 domain constraint is either too strict or uses
a different class hierarchy path than the entity's declared types.  Two
factors mitigate this loss.  First, the agent can still discover a filtered
property if another entity in the same interaction turn exposes it without a
conflict.  Second, SAGA retains all properties whose schema is absent
(see \emph{Type-Constrained Grounding} in the main paper), so the filter
conservatively passes anything it cannot classify. The net gain indicates that
removing clearly wrong candidates outweighs the occasional loss of a correct one.


\section{Boolean Question Analysis}
\label{app:boolean}

Table~\ref{tab:bool_full} reports Boolean results for four Wikidata
datasets, extending the LC-QuAD 2.0 breakdown in the main paper.  SAGA improves Boolean F1 on every
dataset, with the largest gain on QALD-9-plus (33.3\% $\to$ 100.0\%, $+$66.7
points), followed by QALD-7 (25.0\% $\to$ 62.5\%, $+$37.5 points), the
already-reported LC-QuAD 2.0 gain (25.6\% $\to$ 55.6\%, $+$30.0 points), and
QALD-10 (27.9\% $\to$ 52.5\%, $+$24.6 points).  The 100\% Boolean F1 on QALD-9-plus
reflects that SAGA correctly identifies the answer type for all three Boolean
questions in that test set.

\begin{table}[h]
\centering
\caption{Boolean F1 (\%) by dataset.
\#Bool is the number of Boolean questions in the test set.}
\label{tab:bool_full}
\setlength{\tabcolsep}{5pt}
\small
\begin{tabular}{lrcc}
\toprule
Dataset & \#Bool & SPINACH & SAGA \\
\midrule
QALD-7       &  8 & 25.0  & \textbf{62.5} \\
QALD-9+      &  3 & 33.3  & \textbf{100.0} \\
QALD-10      & 61 & 27.9  & \textbf{52.5} \\
LC-QuAD~2.0  & 90 & 25.6   & \textbf{55.6} \\
\bottomrule
\end{tabular}
\end{table}

The answer-type prior (see \emph{Agent Loop} in the main paper) is the primary driver of
the Boolean improvement: it biases the LLM toward ASK queries before any
entity lookup is issued.  The high accuracy of this prior (100\% on QALD-7,
QALD-9-plus, and QALD-10 Boolean questions; 91\% on LC-QuAD 2.0) confirms
that the expected answer type can be reliably inferred from the question
surface form for Boolean queries.


\section{Additional Qualitative Examples}
\label{app:qualitative}

This appendix consolidates all qualitative results and case studies from the
paper.  It provides four Type-A case studies (SPINACH wrong, SAGA correct)
that illustrate different manifestations of type-blind grounding, a compact
set of additional prediction-level snapshots drawn directly from the saved
result files, and a structured failure analysis of several Type-B cases (SAGA
also fails).

\subsection{Entity-Type Recovery: ``Was U.S.\ president Jackson involved in a war?''}
\label{app:jackson-case}

\begin{figure}[h]
\centering
\small
\setlength{\fboxsep}{5pt}
\fbox{\begin{minipage}{0.96\columnwidth}
\textbf{Q}: ``Was U.S.\ president Jackson involved in a war?'' \quad (QALD-7)\\[3pt]
\noindent\rule{\linewidth}{0.4pt}\\[2pt]
\begin{tabular}{@{}p{0.44\linewidth}|p{0.46\linewidth}@{}}
\textbf{SPINACH (type-blind)} &
\textbf{SAGA (type-aware)} \\[3pt]
\textit{Entity search:}\newline
``Jackson'' $\to$ \texttt{Q212}\newline
{\scriptsize (United States; types: sovereign state, territory)} &
\textit{Entity search:}\newline
``Jackson'' $\to$ \texttt{Q212}\newline
{\scriptsize (United States; types: sovereign state, territory)} \\[4pt]
\textit{Neighborhood} (131 props, no type info):\newline
\texttt{P31}: instance of\newline
\texttt{P17}: country\newline
\texttt{P607}: conflict \quad $\leftarrow$ \textit{LLM picks}\newline
$\vdots$ &
\textit{Schema-compact neighborhood} (47 props removed; 84 retained):\newline
\texttt{P31}: instance of [\textit{entity}]\newline
\texttt{P17}: country [\textit{location}]\newline
\textit{P607 filtered} [\textit{domain: \{char., battle\} $\not\ni$ state}]\newline
$\vdots$ \\[4pt]
\textit{Generated SPARQL:}\newline
\texttt{\small ASK WHERE \{}\newline
\texttt{\small \;wd:Q212 wdt:P607 ?w .}\newline
\texttt{\small \}} &
\textit{P607 absent $\Rightarrow$ agent retries:}\newline
``Jackson'' $\to$ \texttt{Q11817}\newline
{\scriptsize (Andrew Jackson; type: human)}\newline
\texttt{P607} passes [\textit{domain $\ni$ human} $\checkmark$]\newline
\textit{Generated SPARQL:}\newline
\texttt{\small SELECT ?c WHERE \{}\newline
\texttt{\small \;wd:Q11817 wdt:P607 ?c .}\newline
\texttt{\small \}} \\[4pt]
$\times$ Wrong: Q212 $\neq$ Jackson &
$\checkmark$ Correct \\
\end{tabular}
\end{minipage}}
\caption{Type-blind grounding (SPINACH, left) resolves ``Jackson'' to
\texttt{Q212} (United States) and selects \texttt{P607} without checking
domain compatibility.  SAGA (right) removes \texttt{P607} from Q212's
131-property neighborhood because its domain \{character, battle, sabotage\}
conflicts with Q212's type \emph{sovereign state} (47 of 131 properties are
removed in total).  The absent \texttt{P607} signals a wrong entity; the agent
retries and finds \texttt{Q11817} (Andrew Jackson), where \texttt{P607}
satisfies the domain constraint.}
\label{fig:jackson}
\end{figure}

\subsection{Entity Type Mismatch: ``Who is the author of WikiLeaks?''}
\label{app:wikileaks-case}

\begin{figure}[h]
\centering
\small
\setlength{\fboxsep}{5pt}
\fbox{\begin{minipage}{0.96\columnwidth}
\textbf{Q}: ``Who is the author of WikiLeaks?'' \quad (QALD-9-plus)\\[3pt]
\noindent\rule{\linewidth}{0.4pt}\\[2pt]
\begin{tabular}{@{}p{0.44\linewidth}|p{0.46\linewidth}@{}}
\textbf{SPINACH (type-blind)} &
\textbf{SAGA (type-aware)} \\[3pt]
\textit{Entity search:}\newline
``WikiLeaks'' $\to$ \texttt{Q23393}\newline
{\scriptsize (types: excerpt, logion, \newline Christian prayer)} &
\textit{Entity search:}\newline
``WikiLeaks'' $\to$ \texttt{Q23393}\newline
{\scriptsize (types: excerpt, logion, \newline Christian prayer)} \\[4pt]
\textit{Neighborhood} (22 props, no type info):\newline
\texttt{P18}: image\newline
\texttt{P112}: founder $\leftarrow$ \textit{LLM picks}\newline
\texttt{P571}: inception\newline
$\vdots$ &
\textit{Schema-compact} (10 of 22 removed):\newline
\texttt{P18}: image [\textit{work}]\newline
\textit{P112 filtered} [\textit{domain:\{community,\newline project\} $\not\ni$ prayer}]\newline
$\vdots$\newline
$\Rightarrow$ \textit{agent retries entity search}\newline
``WikiLeaks'' $\to$ \texttt{Q359}\newline
{\scriptsize (WikiLeaks; type: organisation)}\newline
\texttt{P112} passes [\textit{domain $\ni$ org.}\ $\checkmark$] \\[4pt]
\textit{Generated SPARQL:}\newline
\texttt{\small SELECT ?f WHERE \{}\newline
\texttt{\small \;wd:Q23393 wdt:P112 ?f\}}\newline
$\times$ Wrong entity (Q23393 is a prayer) &
\textit{Generated SPARQL:}\newline
\texttt{\small SELECT ?f WHERE \{}\newline
\texttt{\small \;wd:Q359 wdt:P112 ?f\}}\newline
$\checkmark$ Correct \\
\end{tabular}
\end{minipage}}
\caption{SPINACH resolves ``WikiLeaks'' to \texttt{Q23393}---a
\emph{logion}/\emph{Christian prayer}---whose 22-property neighborhood
contains \texttt{P112} (founder) without any type annotation, so the LLM
selects it.  SAGA removes \texttt{P112} from Q23393's neighborhood because
its domain \{community, project, award\} is incompatible with \emph{excerpt}
and \emph{logion} (10 of 22 properties removed).  The absent founder property
signals the wrong entity; the agent retries and finds \texttt{Q359}
(WikiLeaks, type: organisation), where \texttt{P112} passes the domain check.}
\label{fig:wikileaks}
\end{figure}

\subsection{Tiny-Neighborhood Signal: ``What is the highest place of Karakoram?''}

\begin{figure}[h]
\centering
\small
\setlength{\fboxsep}{5pt}
\fbox{\begin{minipage}{0.96\columnwidth}
\textbf{Q}: ``What is the highest place of Karakoram?'' \quad (QALD-9-plus)\\[3pt]
\noindent\rule{\linewidth}{0.4pt}\\[2pt]
\begin{tabular}{@{}p{0.44\linewidth}|p{0.46\linewidth}@{}}
\textbf{SPINACH (type-blind)} &
\textbf{SAGA (type-aware)} \\[3pt]
\textit{Entity search:}\newline
``Karakoram'' $\to$ \texttt{Q12416}\newline
{\scriptsize (type: Wikimedia location map template)} &
\textit{Entity search:}\newline
``Karakoram'' $\to$ \texttt{Q12416}\newline
{\scriptsize (type: Wikimedia location map template)} \\[4pt]
\textit{Neighborhood} (2 props total):\newline
\texttt{P31}: instance of\newline
\texttt{P610}: highest point $\leftarrow$ \textit{LLM picks}\newline
&
\textit{Schema-compact} (1 of 2 removed):\newline
\texttt{P31}: instance of [\textit{entity}]\newline
\textit{P610 filtered} [\textit{domain:\{region,\newline geographic feature\} $\not\ni$ template}]\newline
\newline
$\Rightarrow$ \textit{only 1 prop retained; agent retries}\newline
``Karakoram'' $\to$ \texttt{Q5469}\newline
{\scriptsize (mountain range; type: geographic feature)}\newline
\texttt{P610} passes [\textit{domain $\ni$ geo.}\ $\checkmark$] \\[4pt]
\textit{Generated SPARQL:}\newline
\texttt{\small SELECT ?peak WHERE \{}\newline
\texttt{\small \;wd:Q12416 wdt:P610 ?peak\}}\newline
$\times$ Q12416 is a template page &
\textit{Generated SPARQL:}\newline
\texttt{\small SELECT ?p WHERE \{}\newline
\texttt{\small \;wd:Q5469 wdt:P610 ?p\}}\newline
$\checkmark$ Correct \\
\end{tabular}
\end{minipage}}
\caption{SPINACH resolves ``Karakoram'' to \texttt{Q12416} (a Wikimedia
location map template), which has only 2 candidate properties; without type
information the LLM selects \texttt{P610} (highest point) anyway.  SAGA's
filter removes \texttt{P610} because its domain \{region, fictional location,
geographical feature\} excludes \emph{Wikimedia template}: with only 1
property remaining in the schema-compact neighborhood, the agent detects the
implausible entity immediately and retries, finding \texttt{Q5469} (Karakoram
mountain range) where \texttt{P610} is domain-compatible.}
\label{fig:karakoram}
\end{figure}

\subsection{Property-Level Type Error: ``In which films directed by Garry Marshall was Julia Roberts starring?''}

\begin{figure}[h]
\centering
\small
\setlength{\fboxsep}{5pt}
\fbox{\begin{minipage}{0.96\columnwidth}
\textbf{Q}: ``In which films directed by Garry Marshall was Julia Roberts starring?''
\quad (QALD-7)\\[3pt]
\noindent\rule{\linewidth}{0.4pt}\\[2pt]
\begin{tabular}{@{}p{0.44\linewidth}|p{0.46\linewidth}@{}}
\textbf{SPINACH (type-blind)} &
\textbf{SAGA (type-aware)} \\[3pt]
\textit{Entity search:}\newline
``Garry Marshall'' $\to$ \texttt{Q115076}\newline
{\scriptsize (types: TV station, state media)}\newline
``Julia Roberts'' $\to$ \texttt{Q1860}\newline
{\scriptsize (types: standard language, literary language)} &
\textit{Entity search:}\newline
``Garry Marshall'' $\to$ \texttt{Q115076}\newline
{\scriptsize (TV station; types incompatible with P57/P161)} \\[4pt]
\textit{Neighborhood of Q115076} (13 props):\newline
\texttt{P57}: director $\leftarrow$ \textit{used}\newline
\texttt{P161}: cast member $\leftarrow$ \textit{used}\newline
$\vdots$\newline
\textit{No type check $\Rightarrow$ LLM uses wrong entities,}\newline
\textit{generates hallucinated query} &
\textit{Schema-compact} (3 of 13 removed):\newline
\textit{P57 filtered} [\textit{domain:\{audiovisual work\}\newline $\not\ni$ TV station}]\newline
\textit{P161 filtered} [\textit{domain:\{work, scene\}\newline $\not\ni$ TV station}]\newline
$\Rightarrow$ \textit{agent retries entity search}\newline
$\to$ \texttt{Q315087} (Garry Marshall, \textit{human})\newline
$\to$ \texttt{Q40523} (Julia Roberts, \textit{human}) \\[4pt]
\textit{Generated SPARQL:}\newline
{\ttfamily\scriptsize SELECT ?dir WHERE \{ ?dir wdt:P106}\newline
{\ttfamily\scriptsize\;wd:Q2526255; wdt:P569 ?bd\}}\newline
{\scriptsize (query about directors born 1920--1930; wrong)}\newline
$\times$ Completely hallucinated &
\textit{Generated SPARQL:}\newline
\texttt{\small SELECT ?f WHERE \{}\newline
\texttt{\small \;?f wdt:P57 wd:Q315087;\newline
\;\;wdt:P161 wd:Q40523\}}\newline
$\checkmark$ Correct \\
\end{tabular}
\end{minipage}}
\caption{SPINACH resolves ``Garry Marshall'' to \texttt{Q115076} (a TV
station) and ``Julia Roberts'' to \texttt{Q1860} (English language), then
applies \texttt{P57} (director) and \texttt{P161} (cast member) without any
type check.  The resulting SPARQL has no connection to the original question.
SAGA filters both \texttt{P57} and \texttt{P161} from Q115076's neighborhood
(3 of 13 removed) because their domains require audiovisual works or scenes,
not a TV station.  The agent retries and correctly resolves both persons,
producing a SPARQL that matches the gold.}
\label{fig:garryjulia}
\end{figure}

\subsection{Additional Prediction Snapshots}
\label{app:qualitative-snapshots}

Unlike the manually inspected figures above, the stored result files do not
preserve full neighborhood renderings for every question.  Table~\ref{tab:qual-snapshots}
therefore reports only query-level differences that are directly observable
from the saved SPINACH \texttt{per\_example.csv} outputs and the matching
SAGA \texttt{predictions.json} files.

\begin{table*}[t]
\centering
\caption{Additional SPINACH-wrong / SAGA-correct snapshots drawn directly from saved predictions.}
\label{tab:qual-snapshots}
\setlength{\tabcolsep}{5pt}
\small
\begin{tabular}{p{0.10\textwidth}p{0.31\textwidth}p{0.26\textwidth}p{0.25\textwidth}}
\toprule
Dataset & Question & SPINACH prediction & SAGA prediction \\
\midrule
QALD-7 &
``Does the Isar flow into a lake?'' &
Uses \texttt{wd:Q4379} for ``Isar'' instead of the river entity and asks for its mouth. &
Uses river \texttt{wd:Q106588} and checks whether its mouth is an instance of \texttt{lake}. \\
QALD-9-plus &
``How many awards has Bertrand Russell?'' &
Counts awards for \texttt{wd:Q131691}, a wrong entity resolution for Bertrand Russell. &
Counts awards for philosopher \texttt{wd:Q33760}, matching the intended entity. \\
QALD-9-plus &
``How many seats does the home stadium of FC Porto have?'' &
Retrieves a stadium and reads \texttt{P1082} (population), a semantically wrong property for seat count. &
Queries stadium capacity with \texttt{P1083} on \texttt{wd:Q271454}. \\
QALD-9-plus &
``What is the revenue of IBM?'' &
Falls onto a generic currency query rather than an IBM revenue statement. &
Uses IBM \texttt{wd:Q37156} and retrieves revenue through statement property \texttt{P2139}. \\
\bottomrule
\end{tabular}
\end{table*}

\subsection{SAGA Failure Cases and Discussion}
\label{app:saga-failures}

SAGA's schema constraints reduce type-incompatible search, but they do not
guarantee correct query composition or predicate interpretation.  Table~\ref{tab:saga-failures}
collects five failures from the saved predictions.  The examples cover
Boolean polarity, missing joins, ontology-level predicate confusion, answer
type propagation, and a controller output that is unrelated to the question.

\begin{table*}[t]
\centering
\caption{Representative SAGA failures (all have execution F1 $=0$).  The
predictions and gold queries are taken from the saved SAGA result files.}
\label{tab:saga-failures}
\setlength{\tabcolsep}{4pt}
\small
\begin{tabular}{p{0.20\textwidth}p{0.27\textwidth}p{0.42\textwidth}}
\toprule
Question & SAGA prediction & Failure observed from the gold comparison \\
\midrule
``Is Frank Herbert still alive?'' (QALD-7) &
\texttt{SELECT} death date with \texttt{wdt:P570} &
The gold query is a Boolean \texttt{ASK} checking that no death date is bound.  SAGA reverses the polarity and returns \texttt{true}, while the gold answer is \texttt{false}. \\
``How many people live in the capital of Australia?'' (QALD-7) &
Returns only the capital via \texttt{wd:Q408 wdt:P36 ?capital} &
The gold query adds the population hop \texttt{?capital wdt:P1082 ?number}; SAGA stops one relation too early and returns Canberra rather than \texttt{381488}. \\
``Was torture the cause of death for Anthony Bourdain?'' (LC-QuAD 2.0) &
Selects \texttt{wdt:P509} (cause of death) &
The gold uses \texttt{wdt:P1196} (manner of death) with a Boolean \texttt{ASK}.  SAGA chooses a related but different Wikidata predicate and produces the wrong Boolean result. \\
``In which city was the president of Montenegro born?'' (QALD-9-plus) &
Returns the president through \texttt{wd:Q236 wdt:P35 ?president} &
The answer requires continuing through \texttt{wdt:P19} and restricting the result to a city.  SAGA identifies the first relation but does not complete the required answer path. \\
``Which actors play in Big Bang Theory?'' (QALD-9-plus) &
Generates an unrelated director--film--award query &
The prediction is disconnected from the question and has F1 $=0$.  The saved prediction alone does not establish whether this arose from context contamination, a failed retry, or controller degeneration. \\
\bottomrule
\end{tabular}
\end{table*}

\paragraph{Discussion.}
The first two examples show that schema-aware grounding does not enforce the
question's logical form.  In the Frank Herbert example, the entity and death
property are both correct, but the requested negation is lost when the agent
switches from an \texttt{ASK} condition to a non-empty \texttt{SELECT}.  In the Australia
example, the first hop is also correct, but the controller terminates before
reaching the requested numeric answer.  These errors suggest that a separate
answer-type and query-completeness check is needed after schema-guided
property selection.

The Anthony Bourdain example exposes a different limitation: domain and range
compatibility cannot by themselves distinguish semantically adjacent
predicates such as ``cause of death'' and ``manner of death.''  Predicate
selection needs finer-grained relation descriptions, demonstrations, or
execution-based verification.  Similarly, the Montenegro example shows that
typing the intermediate entity is not enough when the answer type applies to a
later variable; type constraints should be propagated along the planned path,
not only attached to the first grounded entity.

Finally, the Big Bang Theory example is a controller-level failure rather than
a localized grounding error.  SAGA's schema tools cannot recover when the
controller emits a query for a different task.  A lightweight consistency
guard could compare the generated query's entities, predicates, and answer
shape against the question, then trigger a fresh attempt when the mismatch is
large.  Because the stored result files do not include a causal trace for this
failure, the proposed explanation is a mitigation hypothesis rather than a
confirmed diagnosis.

\subsection{Failure Case: ``Give me all communist countries.''}

\begin{figure}[h]
\centering
\small
\setlength{\fboxsep}{5pt}
\fbox{\begin{minipage}{0.96\columnwidth}
\textbf{Q}: ``Give me all communist countries.'' \quad (QALD-7)\\[3pt]
\noindent\rule{\linewidth}{0.4pt}\\[2pt]
\begin{tabular}{@{}p{0.44\linewidth}|p{0.46\linewidth}@{}}
\textbf{SAGA (P31 available --- normal path)} &
\textbf{SAGA (P31 absent --- fallback)} \\[3pt]
\textit{Entity search:}\newline
answer-class entity $\to$ \texttt{Q}$x$\newline
{\scriptsize (\texttt{P31}: instance-of triples available)}\newline
\textit{Schema-compact neighborhood:}\newline
type-incompatible props filtered\newline
$\Rightarrow$ LLM steered toward\newline
correct class entity (\texttt{Q849866},\newline
communist state) &
\textit{Entity search:}\newline
``communist countries'' $\to$ \texttt{Q3624078}\newline
{\scriptsize (sovereign state; \textbf{no P31 triples in KG})}\newline
\textit{Neighborhood shown:}\newline
all props passed through\newline
(no type info to filter)\newline
$\Rightarrow$ LLM uses \texttt{Q3624078}\newline
(sovereign state) as answer class \\[4pt]
\textit{Generated SPARQL} (hypothetical):\newline
\texttt{\small SELECT ?c WHERE \{}\newline
\texttt{\small \;?c wdt:P31 wd:Q849866 \}}\newline
$\checkmark$ Correct class &
\textit{Generated SPARQL:}\newline
\texttt{\small SELECT ?c WHERE \{}\newline
\texttt{\small \;?c wdt:P122 wd:Q7278 .}\newline
\texttt{\small \;?c wdt:P31 wd:Q3624078 \}}\newline
$\times$ Wrong class (sovereign state\newline
$\neq$ communist state \texttt{Q849866}) \\
\end{tabular}
\end{minipage}}
\caption{The left panel shows the expected SAGA behavior when an entity carries
\texttt{P31} (instance-of) triples: type-incompatible properties are filtered,
steering the agent toward the correct class entity.  The right panel shows
the actual failure: \texttt{Q3624078} (sovereign state) has \emph{no}
\texttt{P31} annotations in Wikidata, so SAGA cannot derive domain constraints
and passes all properties unfiltered.  Without a type signal, the LLM uses the
semantically adjacent but incorrect class \texttt{Q3624078} (sovereign state)
instead of \texttt{Q849866} (communist state).  SAGA's permissive fallback
prevents a crash but cannot supply the missing type information.}
\label{fig:communist}
\end{figure}

\paragraph{Discussion.}
The failure in Figure~\ref{fig:communist} is representative of SAGA's primary
limitation: it can only filter on types that are explicitly declared via
\texttt{P31} (Wikidata's instance-of property).  Class-level entities (e.g.,
\texttt{sovereign state}, \texttt{film}) frequently lack \texttt{P31}
annotations because they \emph{are} the classes, not instances of them.
When such entities appear as answer-class constraints in a SPARQL query,
SAGA's type state is empty and filtering is skipped.  Potential remedies
include using \texttt{P279} (subclass-of) to derive implicit types, building
a complementary class-level schema index, or augmenting type inference with
KG embedding-based entity typing.


\clearpage
\onecolumn

\section{Prompts}
\label{app:prompts}

SAGA issues two LLM calls per question: an \emph{answer-type inference} call
(once, before the agent loop) and a \emph{controller} call (once per agent
step).  When an entity has no type information ($\tau(e)=\emptyset$), a third
\emph{neighborhood pruning} call is used as a fallback
(Section~\ref{app:prompt-prune}); this prompt is inherited unchanged from
SPINACH~\cite{liu2024spinach}.  All templates use Jinja2 syntax:
\texttt{\{\{variable\}\}} is a runtime value substituted at each call and
\texttt{\{\%~\ldots~\%\}} is control flow expanded before the prompt is sent.

\clearpage
\subsection{Answer-Type Prompt}
\label{app:prompt-answertype}

Called once before the first tool call (temperature~$=0$,
\texttt{max\_tokens}~$=100$).  Returns a JSON object with two fields:
\texttt{answer\_type\_label} (a short English type name) and
\texttt{answer\_type\_qid} (a Wikidata QID, or an empty string for scalar
and Boolean answers).  These values seed the persistent type state~$\Theta$
and bias the controller toward ASK queries for Boolean questions
(Appendix~\ref{app:boolean}).  The Freebase variant is identical except that
\texttt{answer\_type\_qid} holds a Freebase type string (e.g.\
\texttt{people.person}) instead of a Wikidata QID.

\begin{lstlisting}[style=promptstyle]
# instruction
Given a natural language question, identify the expected answer type.
Return ONLY a JSON object with two fields:
- "answer_type_label": a concise English label (e.g. "country", "person",
  "date", "number", "organization", "place", "film", "award")
- "answer_type_qid": the most relevant Wikidata QID for this type
  (e.g. "Q6256" for country, "Q5" for human/person, "" if unknown)

If the answer type is unclear or the question asks for a boolean (yes/no),
return {"answer_type_label": "boolean", "answer_type_qid": ""}.
If the answer is a number or quantity, return
{"answer_type_label": "quantity", "answer_type_qid": ""}.

Common mappings:
- person / human -> Q5          - country      -> Q6256
- city           -> Q515        - film / movie -> Q11424
- organization   -> Q43229      - award        -> Q618779
- language       -> Q34770      - university   -> Q3918
- sports team    -> Q12973014

# distillation instruction
Return the answer type JSON.

# input
Question: Who is the president of France?
# output
{"answer_type_label": "person", "answer_type_qid": "Q5"}

# input
Question: In which country was Albert Einstein born?
# output
{"answer_type_label": "country", "answer_type_qid": "Q6256"}

# input
Question: How many films did Stanley Kubrick direct?
# output
{"answer_type_label": "quantity", "answer_type_qid": ""}

# input
Question: {{ question }}
\end{lstlisting}

\clearpage
\subsection{Controller Prompt}
\label{app:prompt-controller}

Called at each agent step (temperature~$=1.0$, top-$p=0.9$,
\texttt{max\_tokens}~$=700$, stop token \texttt{"\textbackslash{}nObservation"}).
Template variables are filled at runtime: \texttt{\{\{answer\_type\}\}} is the
label returned by the answer-type prompt (e.g.\ \textit{person (Q5)});
\texttt{\{\{type\_env\}\}} lists entity-to-type mappings accumulated in~$\Theta$
(at most the five most-recent entries); \texttt{\{\{knowledge\_log\}\}}
summarizes the last six action outcomes; and \texttt{\{\{action\_history\}\}}
is the recent ReAct trajectory (at most the last five steps).
Prior conversation turns are prepended in multi-turn settings.

Lines~20--31 (``Schema-Guided Property Selection'') and lines~33--42
(``Current Type Context'' and ``Knowledge Log'') are the SAGA-specific additions
to the SPINACH controller prompt; the surrounding step-by-step instruction text
is adapted from SPINACH.  The Freebase variant replaces Wikidata QID/PID
conventions with Freebase MID and dotted-property conventions; the structure is
otherwise identical.

After the controller produces free-form text, a second lightweight LLM call
(temperature~$=0$) re-parses the output into a structured JSON object with
fields \texttt{thought}, \texttt{action\_name}, and \texttt{action\_argument}.

\begin{lstlisting}[style=promptstyle]
# instruction
Your task is to write a Wikidata SPARQL query to answer the given question.
Follow a step-by-step process:

1. Start by constructing very simple fragments of the SPARQL query.
2. Execute each fragment to verify its correctness. Adjust as needed based on
   observations.
3. Confirm all your assumptions about the structure of Wikidata before
   proceeding.
4. Gradually build the complete SPARQL query by adding one piece at a time.
5. Do NOT repeat the same action, as the results will be the same.
6. The question is guaranteed to have an answer in Wikidata, so continue
   until you find it.
7. If the user is asking a True/False question with only one answer, use
   ASK WHERE to fetch a True/False answer at the very end.
8. In the final SPARQL projections, ask for the actual entities whenever
   needed.
9. If the final result was contained in last round's get_wikidata_entry and
   you are ready to stop, use execute_sparql to retrieve that result.

## Schema-Guided Property Selection (SAGA)
When selecting a property (PID) from search_wikidata results:
- Results now include [domain: X | range: Y] type constraints where available.
- Domain = the required type of the subject (the entity on the left).
- Range  = the required type of the object (the entity on the right).
- Prefer properties whose domain matches the known type of your subject entity
  and whose range matches the expected answer type.
- If domain/range info is shown, use it to disambiguate between
  similarly-named properties.
- When get_wikidata_entry is called, entity types are shown -- use these to
  check domain compatibility before committing to a property.
- If execute_sparql returns [BLOCKED], the property domain is incompatible
  with the subject type. Use a different property matching the subject's type.
- If execute_sparql returns an empty result, check the Knowledge Log and try
  a different property or entity.

## Current Type Context
Expected answer type: {{ answer_type }}

Known entity types (from explored entities):
{{ type_env }}

## Knowledge Log (recent steps)
{{ knowledge_log }}
([ok] = found results; [fail] Empty/BLOCKED = failed;
 [info] = entity type discovered; avoid repeating failed approaches)

Form exactly one "Thought" and perform exactly one "Action", then wait for
the "Observation".

Possible actions are:

- get_wikidata_entry(QID): Retrieves all outgoing edges (linked entities,
  properties, and qualifiers) of a Wikidata entity using its QID. Also
  reports the entity's types.
- search_wikidata(string): Searches Wikidata for entities or properties
  matching the given string. Property results include type constraints
  [domain: ... | range: ...] when available.
- get_property_examples(PID): Provides a few examples demonstrating the use
  of the specified property (PID) in Wikidata.
- execute_sparql(SPARQL): Runs a SPARQL query on Wikidata and returns a
  truncated result set. If the result is empty with a [WARNING], check
  property type compatibility.
- stop(): Marks the last executed SPARQL query as the final answer and ends.

# distillation instruction
Think and perform the next action.

# input
[Prior conversation turns prepended here in multi-turn settings;
 each turn renders as "User Question: ...\n<action history>\n--"]

User Question: {{ question }}
{{ action_history }}

Output one "Thought" and one "Action":
\end{lstlisting}

\clearpage
\subsection{Fallback Neighborhood Pruning Prompt}
\label{app:prompt-prune}

When no type information is available for an entity ($\tau(e)=\emptyset$), SAGA
falls back to the LLM-based pruning prompt inherited from
SPINACH~\cite{liu2024spinach}.  The prompt is given the raw Wikidata JSON
neighborhood and the current question and asks the LLM to select the relevant
subset.  The two-shot version is shown below (few-shot examples abbreviated
for space; the full examples are in the released code).

\begin{lstlisting}[style=promptstyle]
# instruction
At each turn, you are given a Wikidata entry and a question.
You want to write a SPARQL query that answers the question.
As the first step, remove the parts of the Wikidata entry that could not
be potentially helpful when writing the SPARQL.
The output should be a JSON object containing part of the Wikidata entry.

# distillation instruction
Prune this Wikidata entry.

# input
Wikidata entry for OneRepublic (Q1438730, 'OneRepublic' is an American pop
rock band formed in Colorado Springs, Colorado, in 2002):
{
  "instance of (P31)": "musical group (Q215380)",
  "genre (P136)": ["pop rock (Q484641)", "alternative rock (Q11366)", ...],
  "record label (P264)": "Columbia Records (Q183387)",
  "discography (P358)": "OneRepublic discography (Q935670)",
  ...
}
Question: "What is the title of the second single on OneRepublic's
           third album Native?"
# output
{
  "instance of (P31)": "musical group (Q215380)",
  "record label (P264)": "Columbia Records (Q183387)",
  "discography (P358)": "OneRepublic discography (Q935670)"
}

# input
Wikidata entry for Barack Obama Sr. (Q649593, economist and father of
Barack Obama Jr.):
{
  "educated at (P69)": ["Maseno School (Q6782972)",
                         "University of Hawaii (Q217439)",
                         "Harvard University (Q13371)"],
  "occupation (P106)": "economist (Q188094)",
  ...
}
Question: "where did barack obama sr. attend school?"
# output
{
  "educated at (P69)": ["Maseno School (Q6782972)",
                         "University of Hawaii (Q217439)",
                         "Harvard University (Q13371)"]
}

# input
Wikidata entry for "{{ entity_and_description }}":
{{ outgoing_edges }}

[Context from prior turns if in multi-turn setting]

Current Question: "{{ question }}"
\end{lstlisting}

\end{document}